\documentclass[journal]{IEEEtran}

\usepackage{cite}
\usepackage{amsmath}
\usepackage{algorithmic}
\usepackage{url}
\usepackage{graphicx}
\usepackage{array}
\usepackage{amssymb}
\usepackage{multirow}
\usepackage[caption=false,font=footnotesize]{subfig}
\usepackage[breaklinks=true,letterpaper=true,bookmarks=false]{hyperref}
\usepackage{amsthm}
\usepackage{dsfont}
\usepackage{booktabs}
\usepackage{xcolor}
\usepackage[T1]{fontenc}
\usepackage{algorithmic}
\usepackage[ruled, vlined]{algorithm2e}
\usepackage{graphicx}
\usepackage{subfig}
\usepackage{fancyhdr}

\SetKwInput{KwInput}{Input}             
\SetKwInput{KwOutput}{Output}
\graphicspath{{fig/}}
\def\eg{\emph{e.g.}}

\newcommand{\etal}{\textit{et al}.}
\begin{document}

\title{Saliency Guided Inter- and Intra-Class Relation Constraints for Weakly Supervised Semantic Segmentation}

\author{
	Tao~Chen,
	Yazhou~Yao,
	Lei~Zhang,
	Qiong~Wang*,
	Guo-Sen~Xie*
	and~Fumin~Shen
	
	\thanks{\copyright~2022 IEEE. Personal use of this material is permitted. Permission from IEEE must be
		obtained for all other uses, in any current or future media, including
		reprinting/republishing this material for advertising or promotional purposes, creating new
		collective works, for resale or redistribution to servers or lists, or reuse of any copyrighted
		component of this work in other works.}

}

\markboth{}
{Shell \MakeLowercase{\textit{et al.}}: I2CRC}

\maketitle

\renewcommand{\headrulewidth}{0mm}

\begin{abstract}

 Weakly supervised semantic segmentation with only image-level labels aims to reduce annotation costs for the segmentation task. Existing approaches generally leverage class activation maps (CAMs) to locate the object regions for pseudo label generation. However, CAMs can only discover the most discriminative parts of objects, thus leading to inferior pixel-level pseudo labels. To address this issue, we propose a saliency guided \textbf{I}nter- and \textbf{I}ntra-\textbf{C}lass \textbf{R}elation \textbf{C}onstrained (I$^2$CRC) framework to assist the expansion of the activated object regions in CAMs. Specifically, we propose a saliency guided class-agnostic distance module to pull the intra-category features closer by aligning features to their class prototypes. Further, we propose a class-specific distance module to push the inter-class features apart and encourage the object region to have a higher activation than the background. Besides strengthening the capability of the classification network to activate more integral object regions in CAMs, we also introduce an object guided label refinement module to take a full use of both the segmentation prediction and the initial labels for obtaining superior pseudo-labels. Extensive experiments on PASCAL VOC 2012 and COCO datasets demonstrate well the effectiveness of I$^2$CRC over other state-of-the-art counterparts. The source codes, models, and data have been made available at \url{https://github.com/NUST-Machine-Intelligence-Laboratory/I2CRC}. 

\end{abstract}


\ifCLASSOPTIONpeerreview
	\begin{center} \bfseries EDICS Category: 3-BBND \end{center}
\fi
\IEEEpeerreviewmaketitle

\section{Introduction}

\IEEEPARstart{S}{emantic} segmentation with the goal to label each pixel of an image is a fundamental task in computer vision. With the recent development of deep neural networks, semantic segmentation has achieved remarkable progress. However, the training of a segmentation network requires a large-scale dataset annotated with pixel-level labels \cite{wang2019learning,kang2018depth,everingham2010pascal,lin2014microsoft,chen2021enhanced}. Obtaining such a training dataset is quite labor-intensive and time-consuming. For example, it takes about 90 minutes to annotate each image in the Cityscapes dataset \cite{cordts2016cityscapes}. Therefore, researchers recently turn to weakly supervised learning to alleviate the burden of collecting dense annotations. Various types of weaker labels have been explored to provide supervision for the segmentation task in a weakly supervised setting, \eg, image-level class labels \cite{kolesnikov2016seed,wei2016stc,hong2017weakly,chaudhry2017discovering,huang2018weakly,ahn2018learning,wei2018revisiting,jiang2019integral}, bounding boxes \cite{dai2015boxsup,khoreva2017simple,song2019box}, scribbles \cite{lin2016scribblesup,vernaza2017learning}, and points \cite{bearman2016s}. Since labeling image-level labels takes only one second per class \cite{bearman2016s}, collecting class labels becomes the cheapest and most popular option. In this paper, we aim to address weakly supervised semantic segmentation (WSSS) with the supervision of image-level class labels.

\begin{figure}
	\centering
	\includegraphics[width=\linewidth]{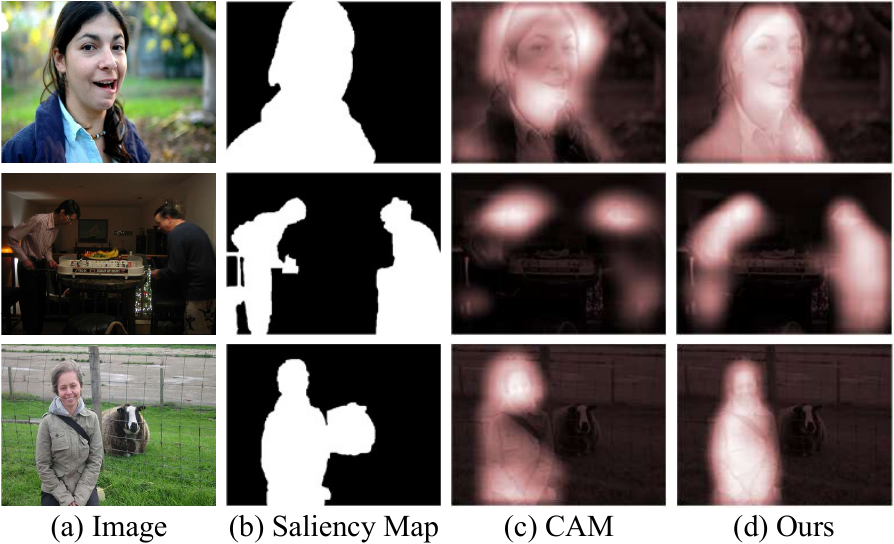}
	\caption{Visual comparisons of localization maps produced by CAM \cite{zhou2016learning} and ours. (a) Input image. (b) Saliency map. (c) Localization maps produced by CAM only identify the most discriminative part of the object, \eg, the head of a person. (d) Localization maps produced by ours can discover more compact and integral object regions. Best viewed in color.}
	\label{fig_moti}
\end{figure}

Though image-level labels can indicate the existence of a specific category of objects, they do not provide any clue about their locations or boundaries. Therefore, it is difficult to directly use them to train segmentation networks. With only image-level class labels, existing works typically train a classification network and rely on class activation maps (CAMs) \cite{zhou2016learning,selvaraju2017grad} to generate pseudo labels for learning a fully supervised semantic segmentation model. However, as illustrated in Fig.~\ref{fig_moti}, CAMs can only locate the most discriminative parts of objects. The small and sparse activated regions bring difficulties in obtaining high-quality pixel-level labels. Therefore, recent research for the WSSS task mainly focuses on enlarging the activated areas in CAMs to cover the entire objects. For example, Jiang \etal~\cite{jiang2019integral} propose an online attention accumulation strategy to gradually identify the integral object regions as the training goes. To further strengthen the lower attention values of target object regions, they also train an integral attention model regarding the cumulative attention maps as supervision. However, when expanding the most discriminative parts to cover more object regions, the background areas around the objects will also be activated inevitably. Therefore, saliency maps \cite{hou2017deeply} have been widely adopted in the WSSS task to provide background clues when producing pseudo labels with CAMs. However, the current usage of saliency maps during the label generation step only provides boundary information of the conspicuous area rather than objects \cite{lee2019ficklenet,jiang2019integral,fan2020learning,fan2020employing,sun2020mining}. In other words, the saliency maps cannot help separate objects of different categories within the salient area and identify the boundaries of objects outside the salient region. Therefore, it is crucial to exploit the potential of saliency maps further to assist the expansion of the activated object regions in CAMs during training while helping regress the integral and compact object area. 

In this paper, we propose a saliency guided inter- and intra-class relation constrained (I$^2$CRC) framework for tackling weakly supervised semantic segmentation. We focus on exploiting the potential of saliency maps while training the classification network to activate more compact and integral object regions in CAMs. Specifically, we propose a saliency guided class-agnostic distance module to explicitly enforce the network to learn consistent and compact feature representations within the salient area. We assume that for the simple images containing only a single category of object(s), as illustrated in the first row of Fig.~\ref{fig_moti}, their saliency maps can be approximated as the ground truth mask for the objects existing in the image. Therefore, with the class-agnostic saliency mask, we can extract the class prototype with masked averaging pooling. We then propose a pixel-level class-agnostic distance (CAD) loss to align salient region features to the object class prototype, which explicitly minimizes the intra-category feature variance. This encourages the classification network to activate more object regions, helping mine the integral objects with CAMs. To prevent the network from learning a tricky position offset for the discriminative region activation to circumvent the intra-category relation constraint, we also apply our saliency guided class-agnostic distance module to the background region with an inversed saliency map. Encouraging the learning of more compact and consistent background features can also help reduce the false activation of irrelevant background regions. To further encourage the object region to have higher activation than the background, we propose a class-specific distance module to push the inter-class features apart. With image-level class labels, we design a prototype-based class-specific distance (CSD) loss to maximize the inter-category feature variance. Echoing with the expectation of higher object activation in CAMs, we explicitly encourage the class-specific object prototype to have a larger value than the background prototype. Note that both CAD loss and CSD loss are only applied to the above mentioned simple images. For complex images having two or more categories of objects, as illustrated in the second and third rows of Fig.~\ref{fig_moti}, the approximation of the saliency map and object mask does not hold. Though inter- and intra-class relation constraints are only applied to simple images, the typical classification loss is applied to both simple and complex images, which endows the network with the ability to distinguish multiple categories in complex images. As shown in Fig.~\ref{fig_moti}, our method can effectively activate more compact and integral object regions for both simple and complex images.

After training a segmentation network with pseudo labels, it is natural to think about leveraging the network prediction to help retrain a more robust segmentation model. We notice that the network predictions and the initial pseudo labels obtained from CAMs can be complementary. Though initial pseudo labels suffer from the loss of boundary information, benefiting from the localization ability of CAMs, they can identify more object instances than the network predictions. In contrast, if the network predictions detect the object instances successfully, they can provide more accurate boundary information. Therefore, we devise an object guided label refinement module to take a full use of the segmentation predictions and initial pseudo labels for deriving higher quality labels. Specifically, we take the segmentation predictions as the basis and change part of their background to object or unreliable labels under the guidance of initial pseudo labels. With the image-level annotations, we also adjust the pixel labels that are predicted as the categories that should not exist in the image and further mine the missed classes in the network predictions.

Our contributions can be summarized as follows:
\begin{itemize}
	\item We propose a saliency guided inter- and intra-class relation constrained (I$^2$CRC) framework for weakly supervised semantic segmentation. We propose a saliency guided class-agnostic distance module and a class-specific distance module to drive the intra-category features closer and the inter-category features apart, respectively.
	
	\item We propose an object guided label refinement module to exploit the potentials of segmentation predictions and initial labels for generating high-quality pseudo-labels.
	
	\item Extensive experiments on PASCAL VOC 2012 and COCO datasets demonstrate state-of-the-art results compared to current methods.
	
\end{itemize}

\section{Related Work}
\subsection{Semantic Segmentation}
Semantic segmentation is the task of assigning a semantic label to every pixel in an image. Based on the fully convolutional network (FCN) \cite{long2015fully}, numerous deep methods have been proposed to promote the development of semantic segmentation \cite{ronneberger2015u, badrinarayanan2017segnet,chen2017deeplab,zhao2017pyramid,zhang2018context,liu2019auto,xie2021scale,xie2021few,chen2021semantically}. For example, the early work of SegNet \cite{badrinarayanan2017segnet} proposed a deep convolutional encoder-decoder architecture to recover the spatial resolution of the input image. Then dilated convolution \cite{chen2017deeplab} was proposed to effectively enlarge the receptive field of filters to incorporate larger context without loss of resolution. To produce high-resolution segmentation maps, Lin \etal~ \cite{lin2017refinenet} proposed a generic multi-path refinement network to combine low-resolution semantic features and fine-grained low-level features in a recursive manner. Recently, self-attention was explored in the works of \cite{wang2018non,fu2019dual,huang2019ccnet} to integrate local features with their global dependencies adaptively. While the work of \cite{wang2018non} leveraged non-local neural networks to capture long-range dependencies, DANet \cite{fu2019dual} proposed both position attention module and channel attention module to learn the spatial and channel interdependencies of features. CCNet \cite{huang2019ccnet} proposed a novel criss-cross attention module to capture contextual information on the criss-cross path. By taking a further recurrent operation, each pixel can finally capture the long-range dependencies from all pixels. Pyramid vision transformer \cite{wang2021pyramid} without convolutions was also proposed for semantic segmentation. Progressive shrinking pyramid and spatial-reduction attention were designed to make the pure transformer backbone flexible to learn multi-scale and high-resolution features. 

\begin{figure*}[h]
	\centering
	\includegraphics[width=\linewidth]{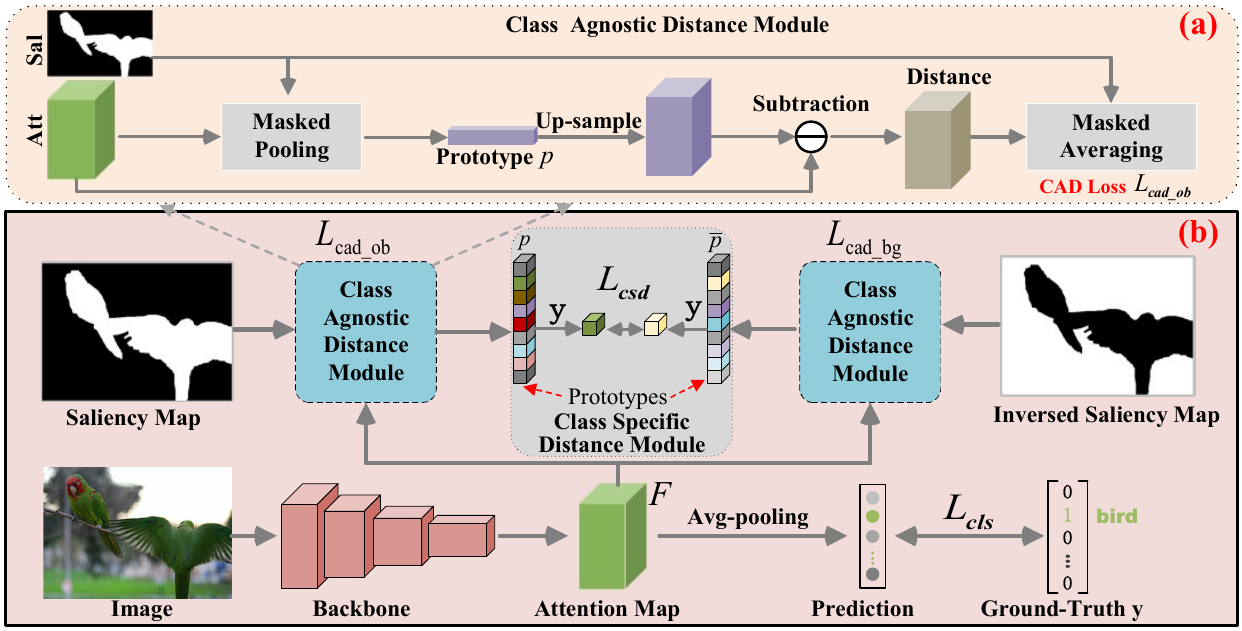}
	\caption{The architecture of I$^2$CRC framework. While training a classification network, we leverage our proposed saliency guided inter- and intra-class relation constraints to expand activated object regions in the localization map. Specifically, a class-agnostic distance module is proposed to align object and background features to their class prototypes.  This helps spread the attention of the most discriminative parts into adjacent non-discriminative object regions and alleviates the irrelevant activation of background regions around objects. Besides, to activate a more compact object region and encourage it to have higher activation than the background, we propose a class-specific distance module to push inter-class features apart. }
	\label{fig_framework}
\end{figure*}

\subsection{Weakly Supervised Semantic Segmentation}
Considering the difficulty of collecting pixel-level labels, researchers resort to address the semantic segmentation task in a weakly supervised setting. Compared to bounding boxes \cite{dai2015boxsup,khoreva2017simple,song2019box}, points \cite{bearman2016s}, and scribbles \cite{lin2016scribblesup,vernaza2017learning}, image-level labels \cite{kolesnikov2016seed,wei2016stc,hong2017weakly,chaudhry2017discovering,huang2018weakly,ahn2018learning,wei2018revisiting,jiang2019integral}have attracted the most attention due to their easy availability. Recently, with the localization ability of CAMs \cite{zhou2016learning}, the performance of weakly supervised semantic segmentation has been significantly improved. With image-level labels, recent methods usually train a classification network and exploit CAMs to discover the most discriminative object regions. However, the object regions highlighted in CAMs are usually small and sparse. Therefore, recent research for the WSSS task mainly focuses on enlarging the activated areas in CAMs to discover entire object regions. For example, the early work of AE-PSL \cite{wei2017object} proposed an adversarial erasing approach to sequentially discover new and complement object regions by erasing the current mined areas. To prohibit attentions from spreading to unexpected background regions, SeeNet \cite{hou2018self} introduced a self-erasing network to promote the quality of object attention. The work of RDC \cite{wei2018revisiting} designed convolutional blocks of different dilated rates to transfer the surrounding discriminative information to non-discriminative object regions. Ahn and Kwak \cite{ahn2018learning} leveraged the initial incomplete discriminative part segmentation to train an AffinityNet for realizing semantic propagation by a random walk with the learned affinities. Sun \etal~\cite{sun2020mining} incorporated two neural co-attentions into the classifier to capture cross-image semantic relations for comprehensive object pattern mining. Li \etal~ \cite{Li2021GroupWiseSM} further proposed a novel group-wise learning framework with a graph neural network operated on a group of images to explore their semantic relations for representation learning.
While these methods expanded the most discriminative parts to cover more object regions, the background areas around the objects would also be activated inevitably. Therefore, they leveraged saliency maps to provide background clues when producing pseudo labels with CAMs. However, the current usage of saliency maps during the label generation step only provides boundary information of the conspicuous area rather than objects. To further exploit the potential of saliency maps for the regression of the integral object region during training, in this work, we propose a saliency guided inter- and intra-class relation constrained framework to assist the expansion of the activated object regions in CAMs.

\section{The Proposed Approach}
In this paper, we propose a saliency guided inter- and intra-class relation constrained (I$^2$CRC) framework for tackling the weakly supervised semantic segmentation task. The framework is illustrated in Fig.~\ref{fig_framework}. Given the image-level weak labels, we train a classification network and leverage the CAMs derived from the attention maps to locate the object regions and generate pseudo labels for the segmentation network training. To enlarge object regions activated in CAMs, we propose a class-agnostic distance module to explicitly minimize the intra-category feature variance. We further maximize the inter-category feature variance with a class-specific distance module to push the object and background features apart. Besides strengthening the classification network's ability to activate more compact and integral object regions in CAMs, we also propose an object guided label refinement module in Section~\ref{sec_refine} to take a full use of the segmentation predictions and initial labels for retraining the segmentation network with high-quality pseudo-labels.

\subsection{CAM Generation}
As in many other weakly supervised semantic segmentation approaches, we locate the object with CAMs, which typically highlight the most discriminative parts of objects. We generate CAMs from the $C$-channel attention map $F$ of the last class-aware convolutional layer, which is proven by \cite{zhang2018adversarial} to be identical to the attention generation process in the original CAMs \cite{zhou2016learning}. Here $C$ is the number of classes. Therefore, each channel of the attention maps corresponds to the CAM for a specific category. The base classification network architecture is illustrated at the bottom of Fig.~\ref{fig_framework} (b).  Finally, global average pooling (GAP) is adopted to get the prediction $q^{c}$ for the $c$-th category. For the training of the classification network, we adopt the multi-label soft margin loss as follows:
\begin{equation}
L_{cls}=-\frac{1}{C} \sum_{c=1}^{C} y^{c} \log \sigma\left(q^{c}\right)+\left(1-y^{c}\right) \log \left[1-\sigma\left(q^{c}\right)\right].
\end{equation}
Here, $\sigma\left(\cdot \right)$ is the sigmoid function. $y^{c}$ is the image-level label for the $c$-th class. Its value is 1 if the class is present in the image; otherwise, its value is 0.

To obtain CAM for each target category $c$, we feed attention map $F^{c}$ into a ReLU layer and then normalize it to range from 0 to 1:
\begin{equation}
A^{c}=\frac{ReLU\left ( F^{c} \right )}{\max\left ( F^{c} \right )}.
\end{equation}
We then discover the discriminative image regions for a particular class following the work of OAA \cite{jiang2019integral}. These regions are further thresholded to generate pixel-level pseudo ground-truths for the segmentation network training.

\subsection{Class-Agnostic Distance Module}
Though CAMs can locate objects in the image, they only activate the most discriminative regions. Besides, without boundary clues, the object activation in CAMs will be spread into the background area as well. These defects of CAMs pose a challenge to the generation of pixel-level pseudo labels for training a segmentation network. Specifically, with the guidance of the saliency maps, we propose a class-agnostic distance module to minimize the intra-category feature variance within the salient region. 

Our proposed class-agnostic distance module is illustrated in Fig.~\ref{fig_framework} (a). For the simple image that contains only a single category of object(s), we approximate the class-agnostic saliency map as its object mask. Then masked average pooling is applied with the object mask to obtain the class prototype vector from features. We up-sample the attention maps $F$ to the same size of the mask $M$, and then the $c$-th element of prototype $p$ is calculated as:
\begin{equation}
p^{c}=\frac{\sum_{i=1,j=1}^{h,w} M_{ij} \cdot F_{ij}^{c}}{\sum_{i=1,j=1}^{h,w} M_{ij}}  .
\label{eq_ptype}
\end{equation}	
Here, $h$ and $w$ are the height and width of the saliency map. We then align object features to the class prototype to encourage the network to learn more compact and consistent feature representations within the object region, leading to more integral object activation in CAMs. We up-sample the prototype vector to the same spatial size of the attention map for element-wise subtraction to obtain the feature distance $D$. We then define a class-agnostic distance loss for the object region with a masked averaging operation as follows:
\begin{equation}
L_{cad\_ob} \!= \! \frac{1}{\sum_{i=1,j=1}^{h',w'} M_{ij}} \! \sum_{i=1,j=1}^{h',w'} \! \left ( M_{ij} \! \cdot \! \frac{1}{C}\sum_{c=1}^{C} \left ( D_{ij}^{c} \right )^{2} \right ) .
\label{eq_cad}
\end{equation}
Here, $h', w'$, and $C$ are the height, width, and channel number of the attention map. The benefit of our proposed class-agnostic distance module is twofold. First, reducing the intra-category feature variance encourages the network to activate the whole object region identified by the saliency area. Second, aligning these features to the prototype will depress peak values of the most discriminative region in CAMs, which forces the network to further activate other less discriminative regions to maintain the network's classification ability. However, with only intra-class relation constraint for object regions, the network might learn a tricky position offset to shift object activation into background regions and impair the localization ability of CAMs. Therefore, as illustrated in Fig.~\ref{fig_framework} (b), we also apply our proposed class-agnostic distance module with the inversed saliency map to encourage the compactness and consistency of background features. Similar to Eq.~(\ref{eq_ptype}), the $c$-th element of background prototype $\bar{p}$ is calculated as follows:
\begin{equation}
\bar{p}^{c}=\frac{\sum_{i=1,j=1}^{h,w} \bar{M}_{ij} \cdot F_{ij}^{c}}{\sum_{i=1,j=1}^{h,w} \bar{M}_{ij}}  .
\label{eq_ptype_bg}
\end{equation}
Here, the inversed saliency map $\bar{M}= 1-M$. After obtaining the background feature distance $\bar{D}$ with its prototype, the class-agnostic distance loss for the background area can be defined similarly to Eq.~(\ref{eq_cad}) as follows:
\begin{equation}
L_{cad\_bg} \!= \!\frac{1}{\sum_{i=1,j=1}^{h',w'} \bar{M}_{ij}} \! \sum_{i=1,j=1}^{h',w'} \!\left ( \bar{M}_{ij} \!\cdot \! \frac{1}{C}\sum_{c=1}^{C} \left ( \bar{D}_{ij}^{c} \right )^{2} \right ) .
\label{eq_cad_bg}
\end{equation}

\subsection{Class-Specific Distance Module}
Through the intra-class relation constraint of the above class-agnostic distance module, we effectively minimize the feature variance inside or outside the salient region. However, the more consistent intra-category features do not guarantee that activation of the object region is higher than the background area. There are two other possible situations: 1) the network generates smooth features and results in similar activation for both object and background regions; 2) the network activates the background region more than the object region. To alleviate the above issue, we propose a class-specific distance module to drive object and background features apart and encourage the object region to have higher activation than the background. As illustrated in Fig.~\ref{fig_framework} (b), after obtaining the class prototypes of object and background from the class-agnostic distance module, we leverage the image-level label $y$ to locate the channel related to the specific class that exists in the image. We then define a class-specific distance loss to explicitly encourage the network to generate higher attention values for the object prototype than the background one as follows:
\begin{equation}
L_{csd}=  y \cdot \bar{p} - y \cdot p  .
\label{eq_csd}
\end{equation}
Here, $p$ and $\bar{p}$ are the object and background prototypes defined in Eq.~(\ref{eq_ptype}) and Eq.~(\ref{eq_ptype_bg}), respectively.

With our proposed saliency guided inter- and intra-class relation constraints, the overall training loss of the classification network is as follows:
\begin{equation}
L=L_{cls} +\lambda _{ob}L_{cad\_ob} + \lambda _{bg}L_{cad\_bg} + \lambda _{csd}L_{csd}.
\label{eq_all}
\end{equation}
Here, $\lambda _{ob}$, $\lambda _{bg}$ and $\lambda _{csd}$ are the hyperparameters that control the relative importance of class-agnostic distance losses for object and background,  and class-specific distance loss.

\subsection{Object Guided Label Refinement}
\label{sec_refine}
After leveraging the pseudo labels derived from CAMs to train a segmentation network, the natural idea is to utilize the segmentation model to generate high-quality pseudo labels. However, directly leveraging the model predictions to retrain the network does not improve performance, indicating the quality of the model prediction is inferior to the initial pseudo labels. However, the segmentation predictions and the initial pseudo labels obtained from CAMs can be complementary. Though initial pseudo labels suffer from the loss of boundary information, benefiting from the localization ability of CAMs, they can identify more object instances than the network predictions. In contrast, the network predictions can provide more accurate boundary information. Therefore, we devise an object guided label refinement module to take advantage of them for generating high-quality labels. First, with image-level label $y_{c}$ that denotes the existence of categories, we can filter out the object labels that should not exist in the segmentation prediction $P$ as follows:
\begin{equation}
P_{ij} = \left\{
\begin{array}{ll}
255  & \text {if}\  P_{ij}=c, y_{c}=0 \\
P_{ij}  & \text { otherwise }
\end{array}.
\right.
\end{equation}
The wrong object labels corrected as 255 will be ignored during training, which will help discard the gradients generated by the misleading information. Then we relabel the prediction's background pixels which are not consistent with the initial pseudo label $L$:
\begin{equation}
P_{ij} = \left\{
\begin{array}{ll}
L_{ij}  & \text {if}\  P_{ij}=0, L_{ij}\neq0 \\
P_{ij}  & \text { otherwise }
\end{array}.
\right.
\end{equation}
This adjustment is based on the observation that only pixels within the salient region or with high activation in CAMs are identified as the non-background area in the initial pseudo label. Therefore, we suppose these pixels are reliable and should not be treated as background. Finally, we further mine the objects that are missed in the network prediction and initial label as follows:
\begin{equation}
P_{ij} = \left\{
\begin{array}{ll}
255 & \text {if}\ P_{ij}=0, \exists \ c, \mathrm{s.t.}\ y_{c}=1, \forall (i,j), P_{ij}\neq c\\
P_{ij} & \text { otherwise }
\end{array}.
\right.
\end{equation}
We traverse all categories in the image-level label and check whether any class of object is missed in $P$. Since we do not have any clue to correct the object pixels currently labeled as background, we set all the background as unreliable labels to be ignored during training. We focus on reducing the false-negative rate of pseudo labels and relying on the self-correction ability of the segmentation network to discover the missed objects.

\section{Experiments}
\subsection{Implementation Details}
For the classification network, we adopt the VGG-16 model \cite{simonyan2014very} as our backbone, which is pre-trained on ImageNet \cite{deng2009imagenet}. Following previous works \cite{jiang2019integral,Li2021GroupWiseSM}, we remove all fully connected layers and add three convolutional layers with 512 channels and kernel size 3 $\times$ 3  on the top of the fully-convolutional backbone. A ReLU layer follows each convolutional layer for nonlinear transformation. Then, a class-aware convolutional layer of $C$ channels with kernel size 1 $\times$ 1 is adopted as the pixel-wise classifier to generate attention maps. The momentum and weight decay of the SGD optimizer are 0.9 and $5 \times 10^{-4}$. The initial learning rate is set to $10^{-3}$ and is divided by 10 after epoch 5 and 10. We train the classification network for 30 epochs with batch size = 5. The saliency maps used in this paper are provided by the work of OAA \cite{jiang2019integral}, which are generated by a pretrained saliency detection model \cite{hou2017deeply}.

For the segmentation network, following \cite{chang2020weakly,zhang2020splitting,fan2020employing,chen2020weakly,zhang2020causal,Li2021GroupWiseSM}, we adopt the DeepLab-v2 \cite{chen2017deeplab} framework. Both the VGG-16 \cite{simonyan2014very} and ResNet-101 \cite{he2016deep} backbones are pre-trained on ImageNet \cite{deng2009imagenet}. We use atrous spatial pyramid pooling (ASPP) \cite{chen2017deeplab} as the final classifier and apply an up-sampling layer along with the softmax output to match the size of the input image. The momentum and weight decay of the SGD optimizer are 0.9 and $10^{-4}$. The initial learning rate is set to $ 2.5 \times 10^{-4}$ and is decreased using polynomial decay with a power of 0.9. The segmentation network is trained for 10,000 iterations with batch size = 10.

\subsection{Datasets and Evaluation Metrics}
Following previous works, we evaluate our approach on the PASCAL VOC 2012 dataset \cite{everingham2010pascal} and COCO dataset \cite{lin2014microsoft}. As the most popular benchmark for WSSS, the PASCAL VOC 2012 dataset contains 21 classes (20 object categories and the background) for semantic segmentation. There are 10,582 training images, which are expanded by \cite{hariharan2011semantic}, 1,449 validation images, and 1,456 test images. COCO dataset is a more challenging benchmark with 80 semantic classes and the background. Following previous works \cite{wang2020weakly,Li2021GroupWiseSM}, we use the default train/val splits (80k images for training and 40k for validation) in the experiment. For all the experiments, we only adopt the image-level class labels for training. Standard mean intersection over union (mIoU) is taken as the evaluation metric for the semantic segmentation task.

\begin{table}[t]
	
	\setlength{\tabcolsep}{2.8mm}
	\renewcommand\arraystretch{1.02}
	\centering
	\caption{Quantitative comparisons to previous state-of-the-art approaches on PASCAL VOC 2012 validation and test set with VGG backbone. I: image-level labels, S: saliency maps. }
	\begin{tabular}{{l}*{4}{c}}
		\toprule
		Methods & Publication & Sup. & Val & Test\\
		\midrule
		STC \cite{wei2016stc}&TPAMI17&I+S&49.8 &51.2\\ 
		AE-PSL \cite{wei2017object} &CVPR17&I+S&55.0 &55.7\\ 
		WebS-i2 \cite{jin2017webly}&CVPR17&I+S&53.4 &55.3\\ 
		Hong \etal~ \cite{hong2017weakly} &CVPR17&I&58.1 &58.7\\ 
		DCSP \cite{chaudhry2017discovering} &BMVC17&I+S&58.6 &59.2\\   
		TPL \cite{kim2017two} &ICCV17&I+S&53.1 &53.8\\ 
		GAIN \cite{li2018tell} &CVPR18&I+S&55.3 &56.8\\ 
		DSRG \cite{huang2018weakly} &CVPR18&I+S&59.0 &60.4\\
		MCOF \cite{wang2018weakly} &CVPR18&I+S&56.2 &57.6\\
		AffinityNet \cite{ahn2018learning} &CVPR18&I&58.4 &60.5\\  
		RDC \cite{wei2018revisiting} &CVPR18&I+S&60.4 &60.8\\
		SeeNet \cite{hou2018self} &NIPS18&I+S&63.1 &62.8\\
		SSNet \cite{zeng2019joint} & ICCV19 &I+S&57.1 &58.6\\
		OAA \cite{jiang2019integral}&ICCV19&I+S&63.1 &62.8\\
		IAL \cite{wang2020weakly} &IJCV20&I+S&62.0 &62.4\\  
		ICD \cite{fan2020learning} &CVPR20&I&64.0 &63.9\\
		BES \cite{chen2020weakly}&ECCV20&I&60.1 &61.1\\ 
		Zhang \etal \cite{zhang2020splitting} &ECCV20&I+S&63.7 &64.5\\  
		MCIS \cite{sun2020mining}&ECCV20&I+S&63.5 &63.6\\ 
		Li \etal~\cite{Li2021GroupWiseSM}  &AAAI21&I+S&63.3 &63.6\\	
		ECS-Net \cite{sun2021ecs} &ICCV21&I&62.1 &63.4\\
		\textbf{I$^2$CRC (Ours)} &-&I+S&\textbf{64.3} &\textbf{65.4}\\	
		\bottomrule			
	\end{tabular}
	\label{tab_vgg}	
\end{table}

\subsection{Comparisons to the State-of-the-arts}
\paragraph{\textbf{Baselines}}
In this part, we compare our proposed method with the following state-of-the-art approaches that leverage image-level labels for weakly supervised semantic segmentation: BFBP \cite{saleh2016built}, SEC \cite{kolesnikov2016seed},  STC \cite{wei2016stc}, AE-PSL \cite{wei2017object}, WebS-i2 \cite{jin2017webly}, Hong \etal~ \cite{hong2017weakly}, DCSP \cite{chaudhry2017discovering}, TPL \cite{kim2017two}, GAIN \cite{li2018tell}, DSRG \cite{huang2018weakly}, MCOF \cite{wang2018weakly}, AffinityNet \cite{ahn2018learning}, RDC \cite{wei2018revisiting}, SeeNet \cite{hou2018self}, SSNet \cite{zeng2019joint}, OAA \cite{jiang2019integral}, IAL \cite{wang2020weakly}, ICD \cite{fan2020learning}, BES \cite{chen2020weakly}, Fan \etal \cite{fan2020employing}, Zhang \etal~ \cite{zhang2020splitting}, MCIS \cite{sun2020mining}, IRN \cite{ahn2019weakly}, FickleNet\cite{lee2019ficklenet}, SSDD \cite{shimoda2019self}, SEAM \cite{wang2020self}, SCE \cite{chang2020weakly}, CONTA \cite{zhang2020causal}, LIID \cite{liu2020leveraging}, Li \etal~\cite{Li2021GroupWiseSM}, NSROM \cite{yao2021non}, ECS-Net \cite{sun2021ecs}.

\begin{figure}[t]
	\begin{center}
		\includegraphics[width=\linewidth]{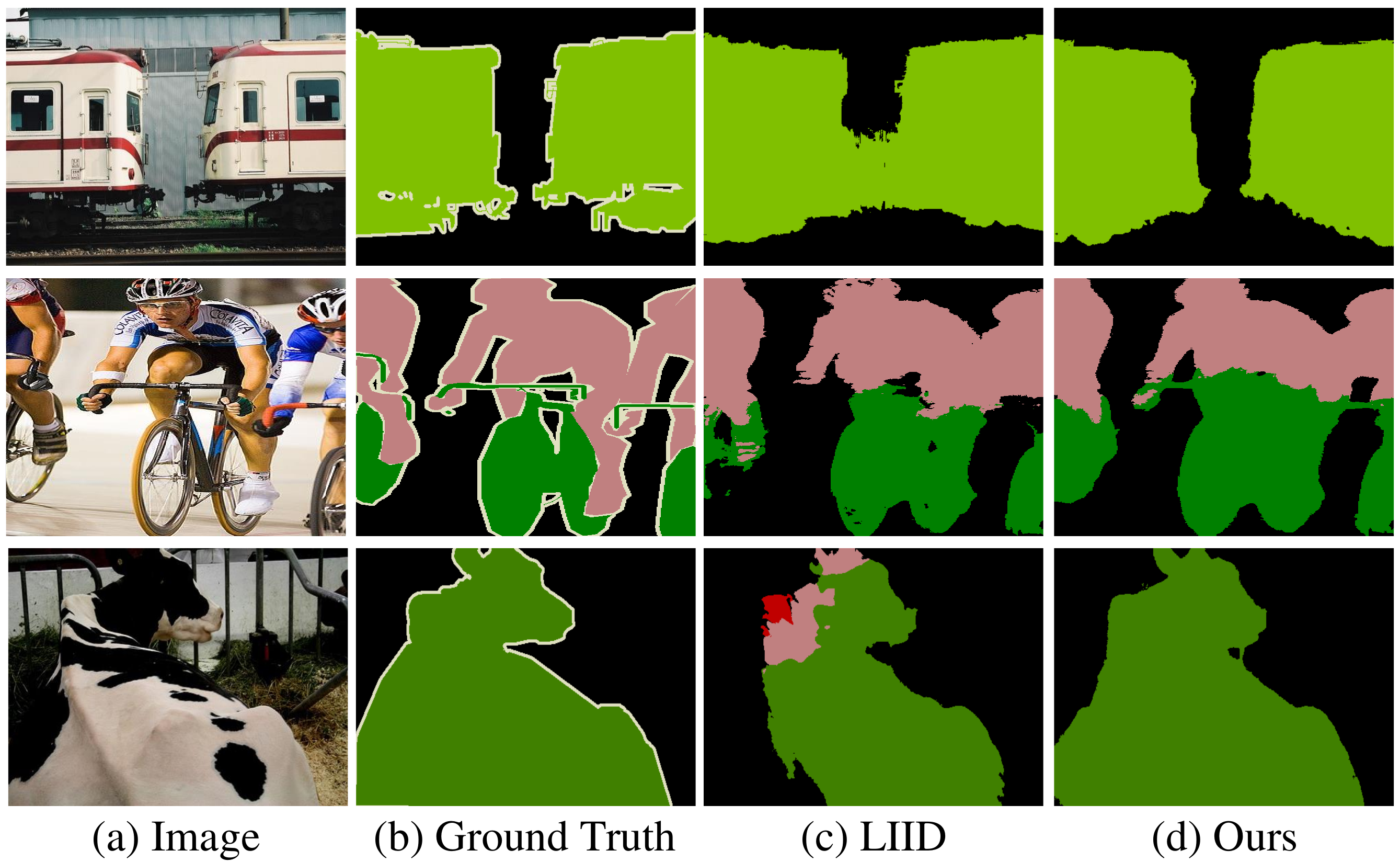}
	\end{center}
	\caption{Visual comparisons with LIID \cite{liu2020leveraging} on PASCAL VOC 2012 validation set. For each (a) image, we show the (b) ground truth, the result of (c) LIID \cite{liu2020leveraging} and (d) our method.  Best viewed in color.}
	\label{sota_compare}
\end{figure}

\paragraph{\textbf{Experimental Results on PASCAL VOC 2012}} We present our results on PASCAL VOC 2012 dataset for the backbone of VGG and ResNet in Table \ref{tab_vgg} and Table \ref{tab_resnet}, respectively. As can be seen, our proposed method achieves better results than other state-of-the-art approaches for both VGG and ResNet backbones. For the VGG backbone, we can get 64.3\% on the validation set and 65.4\% on the test set. Compared with the methods of STC \cite{wei2016stc},  WebS-i2 \cite{jin2017webly} and Hong \etal \cite{hong2017weakly} that utilize additional training images, our method can improve their results by more than +6.7\% mIoU. Our approach outperforms the second-best method proposed by Zhang \etal \cite{zhang2020splitting} by +0.9\% mIoU on the test set. For the more powerful ResNet backbone, our segmentation results reach 69.3\% and 69.5\% on the validation and test set, respectively. Compared to DSRG \cite{huang2018weakly} and CONTA \cite{zhang2020causal} that also leverage the prediction of the model to help train the segmentation network, our approach has +2.8\% performance gain. Compared to the work of LIID \cite{liu2020leveraging} that can tackle both instance and semantic segmentation tasks with weak labels, our proposed approach outperforms it by +2.8\% and +2.0\% on the validation and test set, respectively. The visual comparisons with LIID \cite{liu2020leveraging} is illustrated in Fig.~\ref{sota_compare}. As can be seen, though both LIID \cite{liu2020leveraging} and our proposed method can successfully detect each instance in the image, our proposed method can lead to more accurate segmentation at the boundary. Compared to the current leading approach of NSROM \cite{yao2021non}, our method can outperform it by +1.0\%.

\begin{table}[t]
	
	\setlength{\tabcolsep}{2.8mm}
	\renewcommand\arraystretch{1.02}
	\centering
	\caption{Quantitative comparisons to previous state-of-the-art approaches on PASCAL VOC 2012 validation and test set with ResNet backbone. I: image-level labels, S: saliency maps.}
	\begin{tabular}{{l}*{4}{c}}
		\toprule
		Methods & Publication & Sup. & Val & Test\\
		\midrule
		DCSP \cite{chaudhry2017discovering} &BMVC17&I+S&60.8 &61.9\\
		DSRG \cite{huang2018weakly} &CVPR18&I+S&61.4 &63.2\\ 
		MCOF \cite{wang2018weakly} &CVPR18&I+S&60.3 &61.2\\ 
		AffinityNet \cite{ahn2018learning} &CVPR18&I&61.7 &63.7\\
		SeeNet \cite{hou2018self} &NIPS18&I+S&63.1 &62.8\\
		IRN \cite{ahn2019weakly} &CVPR19&I&63.5 &64.8\\ 
		FickleNet\cite{lee2019ficklenet} &CVPR19&I+S&64.9 &65.3\\
		SSNet \cite{zeng2019joint} & ICCV19 &I+S&63.3 &64.3\\
		OAA \cite{jiang2019integral}&ICCV19&I+S&65.2 &66.4\\
		SSDD \cite{shimoda2019self}&ICCV19&I+S&64.9 &65.5\\   
		IAL \cite{wang2020weakly} &IJCV20&I+S&64.3 &65.4\\
		SEAM \cite{wang2020self} &CVPR20&I+S&64.5 &65.7\\
		SCE \cite{chang2020weakly} &CVPR20&I&66.1 &65.9\\
		ICD \cite{fan2020learning} &CVPR20&I+S&67.8 &68.0\\
		Zhang \etal~ \cite{zhang2020splitting} &ECCV20&I+S&66.6 &66.7\\
		Fan \etal \cite{fan2020employing} &ECCV20&I+S&67.2 &66.7\\
		MCIS \cite{sun2020mining} &ECCV20&I+S&66.2 &66.9\\
		BES \cite{chen2020weakly}&ECCV20&I&65.7 &66.6\\  
		CONTA \cite{zhang2020causal}&NIPS20&I&66.1 &66.7\\
		LIID \cite{liu2020leveraging}&TPAMI20&I&66.5&67.5\\
		Li \etal~ \cite{Li2021GroupWiseSM} &AAAI21&I+S&68.2 &68.5\\	
		NSROM \cite{yao2021non} &CVPR21&I+S&68.3 &68.5\\
		ECS-Net \cite{sun2021ecs} &ICCV21&I&66.6 &67.6\\
		\textbf{I$^2$CRC (Ours)} &-&I+S&\textbf{69.3} &\textbf{69.5}\\		
		\bottomrule	
	\end{tabular}
	
	\label{tab_resnet}	
\end{table}

\begin{figure*}[t]
	\begin{center}
		\includegraphics[width=\linewidth]{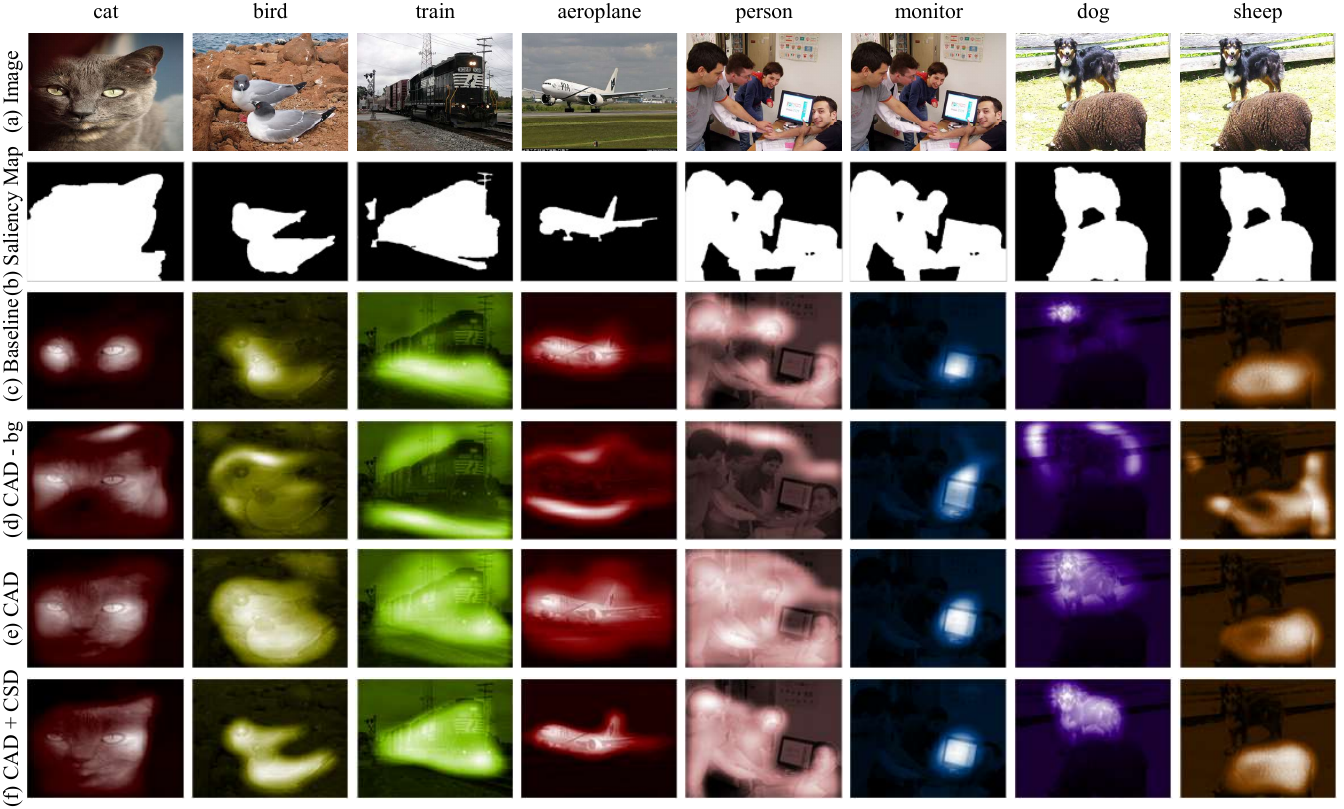}
	\end{center}
	\caption{Visual comparisons about CAMs for the PASCAL VOC 2012 training set. For each (a) image, we show the (b) saliency map, attention maps produced by (c) baseline, (d) CAD - bg, (e) CAD, and (f) CAD + CSD. CAD: Class-Agnostic Distance Module; CSD: Class-Specific Distance Module. The first four columns of images contain objects of only one category, and the other four columns of images contain objects of multiple categories. Best viewed in color.}
	\label{fig_cam}
\end{figure*}

\paragraph{\textbf{Experimental Results on COCO}} To further examine the performance of our proposed approach, we conduct experiments on the more challenging COCO dataset. We present our results for the backbone of VGG in Table \ref{tab_vgg_coco}. As can be seen, our proposed algorithm achieves the best performance of 31.2\% on the validation set. It significantly outperforms the second-best result reported in the work of Li \etal~\cite{Li2021GroupWiseSM} by +2.8\% mIoU, which demonstrates the superiority of our approach.

\begin{table}[t]
	
	\setlength{\tabcolsep}{2.8mm}
	\renewcommand\arraystretch{1.02}
	\centering
	\caption{Quantitative comparisons to previous state-of-the-art approaches on COCO validation set with VGG backbone. I: image-level labels, S: saliency maps.}
	\begin{tabular}{{l}*{3}{c}}
		\toprule
		Methods & Publication & Sup. & Val \\
		\midrule
		BFBP \cite{saleh2016built}  &ECCV16&I&20.4\\ 
		SEC \cite{kolesnikov2016seed} &ECCV16&I&22.4\\ 
		DSRG \cite{huang2018weakly} &CVPR18&I+S&26.0\\ 
		CONTA \cite{zhang2020causal}&NIPS20&I&23.7\\
		IAL \cite{wang2020weakly} &IJCV20&I+S&27.7\\ 
		Li \etal~ \cite{Li2021GroupWiseSM} &AAAI21&I+S&28.4\\
		
		\textbf{I$^2$CRC (Ours)} &-&I+S&\textbf{31.2} \\		
		\bottomrule	
	\end{tabular}
	
	\label{tab_vgg_coco}	
\end{table}

\subsection{Ablation Studies}

\paragraph{\textbf{Element-Wise Component Analysis}}
In this part, we demonstrate the contribution of each component proposed in our approach. Visual comparisons about CAMs are illustrated in Fig.~\ref{fig_cam}. For each image, we show the saliency map, and attention maps produced by (c) baseline: the original method of CAMs  \cite{zhang2018adversarial,zhou2016learning};  (d) CAD - bg: CAD only applied to the object region without the background constraint; (e) CAD: our default setting of applying CAD to both the object and background area; and (f) CAD + CSD: our proposed saliency guided inter- and intra-class relation constraints with both CAD and CSD. As shown in Fig.~\ref{fig_cam} (c), attention maps produced by the baseline method only locate the most discriminative regions of the object, \eg, the eyes of the cat. From Fig.~\ref{fig_cam} (d), we can notice that aligning the object features to their prototype can minimize the intra-category variance. However, as shown in the 3rd to 5th columns, to circumvent the intra-class relation constraint, the network might also learn a tricky position offset to shift activation of the most discriminative region into the background area, which impairs the localization ability of CAMs. From Fig.~\ref{fig_cam} (e), we can notice that applying CAD to both object and background regions can address this activation shift issue and enlarge the activated object region. However, as demonstrated in the 2nd to 5th columns, the network will also expand the object activation to the background when generating smooth features for both object and background area around the object. Finally, with our proposed CSD, we successfully drive object and background features apart and encourage object regions to have higher activation than background. As shown in Fig.~\ref{fig_cam} (f), our approach can effectively activate more compact and integral object regions for both simple and complex images.

\begin{figure*}[t]
	\begin{center}
		\includegraphics[width=\linewidth]{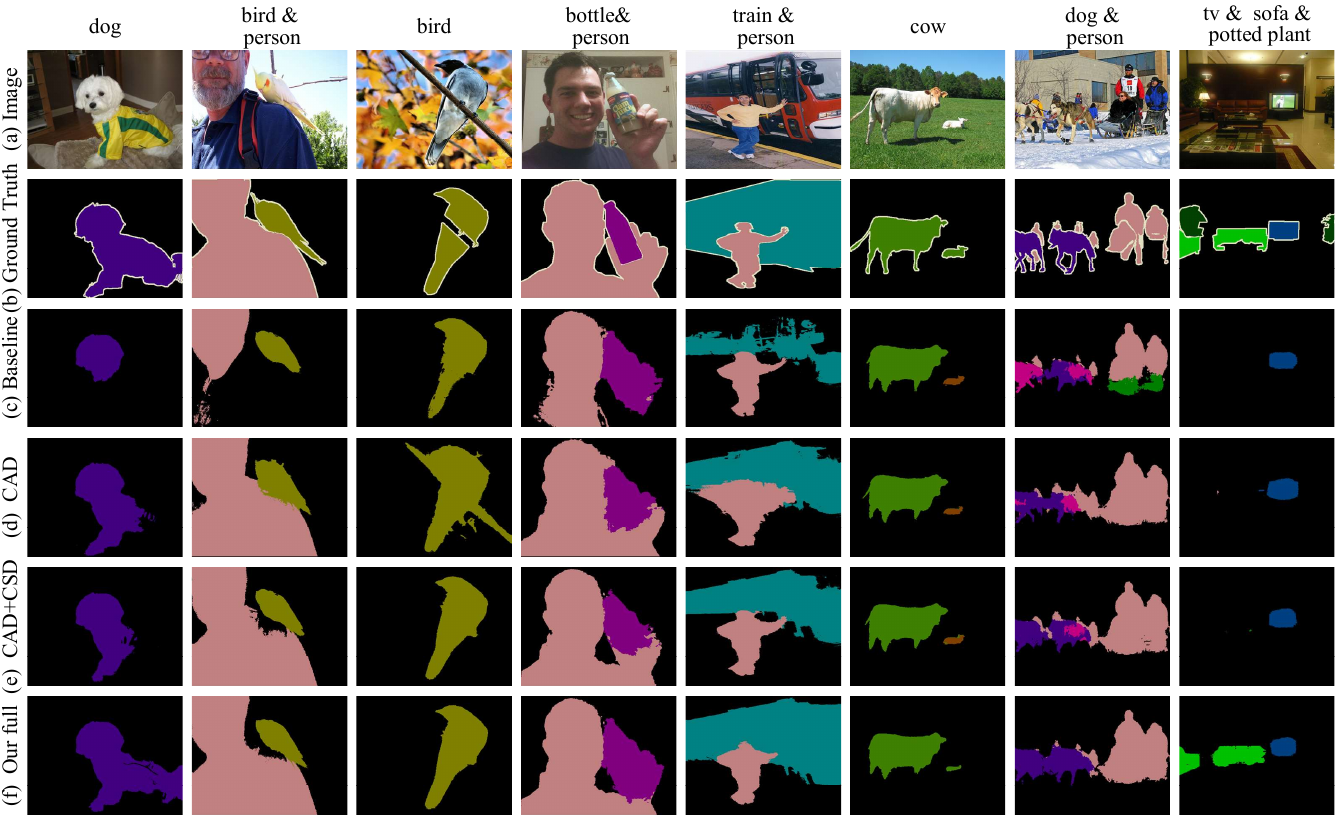}
	\end{center}
	\caption{Example results on PASCAL VOC 2012 validation set. For each (a) image, we show the (b) ground truth, the result of (c) baseline, (d) CAD, (e) CAD + CSD, and (f) our full method. CAD: Class-Agnostic Distance Module; CSD: Class-Specific Distance Module. Best viewed in color.}
	\label{fig_result}
\end{figure*}

The experimental results on the validation set of PASCAL VOC for the ResNet backbone are given in Table \ref{tab_element}. As can be seen, by leveraging our proposed CAD to minimize the intra-category feature variance for both object and background regions, we can improve the baseline result from 63.5\% to 65.4\%. In our experiment, we notice that if CAD is only applied to the object region without the background constraint, the performance  will drop significantly to 61.5\% due to the activation shifting issue illustrated in Fig.~\ref{fig_cam} (d). This highlights the importance of applying CAD to both the object and background area, which maintains the localization ability of CAMs and enlarges the object activation region. By utilizing CSD to drive object and background features apart, we obtain another +2.6\% performance gain and get the mIoU of 68.0\%. We can also note that if CSD is applied alone, the performance drops significantly to 43.3\%. Without the CAD loss, the network cannot benefit from the object boundary information provided by the saliency map and thus leads to inaccurate object activation. Specifically, the network tends to activate more background regions to meet the constraint of CSD loss. This highlights the importance of CAD loss that helps learn compact and consistent features for both object and background. As we can see, after learning a segmentation model with pseudo labels, if we directly retrain the network with its prediction, the performance drops from 68.0\% to 67.2\%. In contrast, with the higher quality labels generated from our proposed object guided label refinement module, retraining the segmentation network can further improve the performance to the mIoU of 69.3\%. 

\begin{table}[t]

	\renewcommand\arraystretch{1.05}
	\caption{Element-wise component analysis on PASCAL VOC 2012 validation set with ResNet backbone. CAD: Class-Agnostic Distance Module; CSD: Class-Specific Distance Module; LR: Object Guided Label Refinement Module.}
	\centering
	\begin{tabular}{{l}*{1}{c}}
		\toprule
		Methods  & Val \\
		\midrule
		baseline &63.5\\
		+ CAD &65.4\\
		+ CSD &43.3\\
		+ CAD + CSD &68.0\\
		\midrule
		+ CAD + CSD + retrain &67.2\\
		+ CAD + CSD + LR &69.3\\		
		\bottomrule	
	\end{tabular}
	\label{tab_element}	
\end{table}

To further verify the effectiveness of our proposed label refinement module, we also evaluate the quality of the pseudo mask on the PASCAL VOC 2012 train set with and without our proposed object guided label refinement module (LR). As can be seen in Table \ref{tab_lr}, without LR, the mIoU of the pseudo mask is only 70.7\%. After object guided label refinement, the quality of the newly generated pseudo mask can reach 76.1\%.

\begin{table}[htb]
	
	\setlength{\tabcolsep}{4mm}
	\renewcommand\arraystretch{1.05}
	\caption{Evaluation of the pseudo mask (Mask) on PASCAL VOC 2012 train set with and without our proposed object guided label refinement module (LR). }
	\centering
	\begin{tabular}{{c}*{1}{c}}
		\toprule
		Methods  & Mask \\
		\midrule
		Ours w/o LR &70.7\\
		Ours w/~~ LR &76.1\\	
		\bottomrule	
	\end{tabular}
	\label{tab_lr}	
\end{table}

Some qualitative segmentation examples on the PASCAL VOC 2012 validation set can be viewed in Fig.~\ref{fig_result}. As shown in the first two columns of Fig.~\ref{fig_result} (d), after enlarging object regions with CAD, we can accordingly segment out more integral objects than the baseline method. Given the more compact object region activated with our saliency guided inter- and intra-class relation constraints (CAD + CSD), as shown in the 3rd to 5th columns of Fig.~\ref{fig_result} (e), we obtain more refined objects from the segmentation map. Finally, exploiting the potential of segmentation prediction and initial labels for generating higher quality pseudo labels, we can correct part of the wrong prediction in images of the 6th and 7th columns and discover the missed sofa in the image of the last column.

\paragraph{\textbf{Comparisons of Pseudo Mask Generated from Saliency Maps and CAMs}} 
In this paper, we assume that for the simple images, their saliency maps can be approximated as the ground truth. Based on the assumption, it seems like there is no need to generate pseudo ground truth from CAM for simple images. In Table \ref{tab_simple}, we compare the quality of pseudo labels generated for simple images of the PASCAL VOC 2012 train set from saliency maps and CAMs. As can be seen, the mIoU between saliency maps for simple images and ground truth is 69.1\%. If generated from CAMs, the quality of pseudo labels for simple images can be improved to 76.6\%. We also compare the performance of using pseudo ground-truth for complex images and saliency maps for simple images to train the segmentation network with our default setting (which trains the network with pseudo labels all generated from CAMs). Note that the results on PASCAL VOC 2012 validation set are reported without our object guided label refinement module. As shown in Table \ref{tab_simple}, compared with using saliency maps for simple images, leveraging pseudo labels all generated from CAMs can obtain 1.6\% performance gain.

\begin{table}[t]
	
	\setlength{\tabcolsep}{4mm}
	\renewcommand\arraystretch{1.05}
	\caption{Evaluation of the pseudo mask generated for simple images (Mask-S) of PASCAL VOC 2012 train set from saliency maps and CAMs. }
	\centering
	\begin{tabular}{{c}*{2}{c}}
		\toprule
		Methods  & Mask-S&Val \\
		\midrule
		Saliency &69.1&66.4\\
		CAM &76.6&68.0\\	
		\bottomrule	
	\end{tabular}
	\label{tab_simple}	
\end{table}

\paragraph{\textbf{The Number of Simple Images}}
In Table \ref{tab_subset}, we show the performance of our method when our proposed losses are applied to the subset of simple images. Note that the results on PASCAL VOC 2012 validation set are reported without our object guided label refinement module. As can be seen, when computing the proposed losses only on 80\%, 50\%, 20\% of the simple images in the PASCAL VOC dataset, the performance drops from 68.0 to 67.8, 67.8, and 67.7, respectively. Reducing the number of simple images only slightly affects the performance, demonstrating the robustness of our proposed losses.

\begin{table}[htb]
	
	\setlength{\tabcolsep}{4mm}
	\renewcommand\arraystretch{1.05}
	\caption{Comparisons of mIoU scores on PASCAL VOC 2012 validation set when our proposed losses are applied on different number of simple images. Note that images are selected randomly. \#Images-S and Proportion-S denote the number and percentage of simple images. }
	\centering
	\begin{tabular}{*{3}{c}}
		\toprule
		\#Images-S &Proportion-S &Val \\
		\midrule
		1349&20\%& 67.7\\
		3373&50\%& 67.8\\
		5397&80\%& 67.8\\
		6746&100\%& 68.0\\	
		\bottomrule	
	\end{tabular}
	\label{tab_subset}	
\end{table}

\begin{table}
	\centering
	\caption{The parameter sensitivity of $\boldsymbol{\lambda _{obj}}$, $\boldsymbol{\lambda _{bg}}$ and $\boldsymbol{\lambda _{csd}}$. Results are reported on PASCAL VOC 2012 validation set with the ResNet backbone.}
	\begin{tabular}{l|l|l|c}
		\toprule
		\multicolumn{3}{c|}{Parameters} & Val \\
		\midrule
		$\lambda _{ob}$ = 0 & $\lambda _{bg}$ = 0 & $\lambda _{csd}$ = 0 & 63.5\\
		\midrule
		$\lambda _{ob}$ = 0.0075 &\multirow{3}*{$\lambda _{bg}$ = 0.025}&\multirow{3}*{$\lambda _{csd}$ = 0.1}&66.8\\
		$\lambda _{ob}$ = 0.015 &&&67.6\\
		$\lambda _{ob}$ = 0.02 &&&66.3\\
		\midrule
		\multirow{4}*{$\lambda _{ob}$ = 0.01}&$\lambda _{bg}$ = 0.02&\multirow{4}*{$\lambda _{csd}$ = 0.1} &67.3\\
		&$\lambda _{bg}$ = 0.03 &&67.6\\
		&$\lambda _{bg}$ = 0.035 &&67.5\\
		&$\lambda _{bg}$ = 0.04 &&65.9\\
		\midrule
		\multirow{3}*{$\lambda _{ob}$ = 0.01}&\multirow{3}*{$\lambda _{bg}$ = 0.025}&$\lambda _{csd}$ =  0.075 &66.9\\
		&&$\lambda _{csd}$ =  0.15 &67.7\\
		&&$\lambda _{csd}$ =  0.2 &67.0\\		
		\midrule
		$\lambda _{ob}$ = 0.01 & $\lambda _{bg}$ = 0.025 & $\lambda _{csd}$ = 0.1 & \textbf{68.0}\\
		\bottomrule	
	\end{tabular}
	
	\label{tab_parameter}	
\end{table}

\paragraph{\textbf{Parameter Analysis}} 
We further conduct parameter analysis on the PASCAL VOC 2012 validation set to verify the effectiveness of the essential modules in our approach. We use ResNet101 as the default backbone. Note that the results are reported without our object guided label refinement module. The baseline result is given in the first row and the performance with default parameters reported in this paper for Eq.~(\ref{eq_all}) is given in the last row of Table \ref{tab_parameter}. 
We first investigate the effect of $\lambda _{ob}$, which controls the relative importance of class-agnostic distance loss for object region. As shown in Table \ref{tab_parameter}, we can get better mIoU score when $\lambda _{ob}$ varies between 0.01 to 0.015. A larger or smaller $\lambda _{ob}$ may not improve the results very much. 
For the parameter that controls the relative importance of class-agnostic distance loss for background, we vary $\lambda _{bg}$ over the range $\left \{0.02, 0.025, 0.03, 0.035, 0.04\right \}$. We can find that the segmentation result is quite stable when $\lambda _{bg}$ varies from 0.02 to 0.035 and a larger $\lambda _{bg}$ deteriorates the performance greatly. 
We further evaluate the impact of the class-specific distance module by comparing the performance with different $\lambda _{csd}$ ranging between $\left \{0.075, 0.1, 0.15, 0.2\right \}$. As shown in Table \ref{tab_parameter}, the mIoU score is better improved when $\lambda _{csd}$ varies between 0.1 to 0.15. A larger or smaller $\lambda _{csd}$ may not improve the results very much. 
On the whole, the results in Table \ref{tab_parameter} are always much better than the baseline result, which demonstrates the effectiveness and robustness of our proposed method. According to Table \ref{tab_parameter}, we finally set $\lambda _{ob}$ = 0.01, $\lambda _{bg}$ = 0.025 and $\lambda _{csd}$ = 0.1.

\section{Conclusions}
In this work, we proposed a saliency guided inter- and intra-class relation constrained (I$^2$CRC) framework for weakly supervised semantic segmentation. Specifically, we proposed a saliency guided class-agnostic distance module to minimize the intra-category feature variance via aligning features to their class prototype. Further, we proposed a class-specific distance module to drive inter-class features apart and encourage the object region to have higher activation than the background. Besides strengthening the classification network's ability to activate more integral object regions in CAMs, we also proposed an object guided label refinement module to further exploit the potential of segmentation predictions and initial labels for generating higher quality pseudo-labels. Extensive experiments on the PASCAL VOC 2012 and COCO datasets demonstrated the superiority of our proposed approach.

\bibliographystyle{IEEEtran}
\bibliography{IEEEreference}

\begin{thebibliography}{10}
\providecommand{\url}[1]{#1}
\csname url@samestyle\endcsname
\providecommand{\newblock}{\relax}
\providecommand{\bibinfo}[2]{#2}
\providecommand{\BIBentrySTDinterwordspacing}{\spaceskip=0pt\relax}
\providecommand{\BIBentryALTinterwordstretchfactor}{4}
\providecommand{\BIBentryALTinterwordspacing}{\spaceskip=\fontdimen2\font plus
\BIBentryALTinterwordstretchfactor\fontdimen3\font minus
  \fontdimen4\font\relax}
\providecommand{\BIBforeignlanguage}[2]{{%
\expandafter\ifx\csname l@#1\endcsname\relax
\typeout{** WARNING: IEEEtran.bst: No hyphenation pattern has been}%
\typeout{** loaded for the language `#1'. Using the pattern for}%
\typeout{** the default language instead.}%
\else
\language=\csname l@#1\endcsname
\fi
#2}}
\providecommand{\BIBdecl}{\relax}
\BIBdecl

\bibitem{wang2019learning}
Q.~Wang, C.~Yuan, and Y.~Liu, ``Learning deep conditional neural network for
  image segmentation,'' vol.~21, no.~7, pp. 1839--1852, 2019.

\bibitem{kang2018depth}
B.~Kang, Y.~Lee, and T.~Q. Nguyen, ``Depth-adaptive deep neural network for
  semantic segmentation,'' vol.~20, no.~9, pp. 2478--2490, 2018.

\bibitem{everingham2010pascal}
M.~Everingham, L.~V. Gool, C.~K. Williams, J.~Winn, and A.~Zisserman, ``The
  pascal visual object classes (voc) challenge,'' in \emph{International
  Journal of Computer Vision}, vol.~88, no.~2, 2010, pp. 303--338.

\bibitem{lin2014microsoft}
T.-Y. Lin, M.~Maire, S.~Belongie, J.~Hays, P.~Perona, D.~Ramanan,
  P.~Doll{\'a}r, and L.~Zitnick, ``Microsoft coco: Common objects in context,''
  in \emph{Proc. European Conference on Computer Vision}, 2014, pp. 740--755.

\bibitem{chen2021enhanced}
T.~Chen, S.~Wang, Q.~Wang, Z.~Zhang, G.~Xie, and Z.~Tang, ``Enhanced feature
  alignment for unsupervised domain adaptation of semantic segmentation,''
  2021.

\bibitem{cordts2016cityscapes}
M.~Cordts, M.~Omran, S.~Ramos, T.~Rehfeld, M.~Enzweiler, R.~Benenson,
  U.~Franke, S.~Roth, and B.~Schiele, ``The cityscapes dataset for semantic
  urban scene understanding,'' in \emph{Proc. IEEE Conference on Computer
  Vision and Pattern Recognition}, 2016, pp. 3213--3223.

\bibitem{kolesnikov2016seed}
A.~Kolesnikov and C.~Lampert, ``Seed, expand and constrain: Three principles
  for weakly-supervised image segmentation,'' in \emph{Proc. European
  Conference on Computer Vision}, 2016, pp. 695--711.

\bibitem{wei2016stc}
Y.~Wei, X.~Liang, Y.~Chen, X.~Shen, M.-M. Cheng, Y.~Zhao, and S.~Yan, ``Stc: A
  simple to complex framework for weakly-supervised semantic segmentation,''
  vol.~39, no.~11, 2016, pp. 2314--2320.

\bibitem{hong2017weakly}
S.~Hong, D.~Yeo, S.~Kwak, H.~Lee, and B.~Han, ``Weakly supervised semantic
  segmentation using web-crawled videos,'' in \emph{Proc. IEEE Conference on
  Computer Vision and Pattern Recognition}, 2017, pp. 7322--7330.

\bibitem{chaudhry2017discovering}
A.~Chaudhry, P.~Dokania, and P.~Torr, ``Discovering class-specific pixels for
  weakly-supervised semantic segmentation,'' in \emph{Proc. British Machine
  Vision Conference}, 2017.

\bibitem{huang2018weakly}
Z.~Huang, X.~Wang, J.~Wang, W.~Liu, and J.~Wang, ``Weakly-supervised semantic
  segmentation network with deep seeded region growing,'' in \emph{Proc. IEEE
  Conference on Computer Vision and Pattern Recognition}, 2018, pp. 7014--7023.

\bibitem{ahn2018learning}
J.~Ahn and S.~Kwak, ``Learning pixel-level semantic affinity with image-level
  supervision for weakly supervised semantic segmentation,'' in \emph{Proc.
  IEEE Conference on Computer Vision and Pattern Recognition}, 2018, pp.
  4981--4990.

\bibitem{wei2018revisiting}
Y.~Wei, H.~Xiao, H.~Shi, Z.~Jie, J.~Feng, and T.~Huang, ``Revisiting dilated
  convolution: A simple approach for weakly-and semi-supervised semantic
  segmentation,'' in \emph{Proc. IEEE Conference on Computer Vision and Pattern
  Recognition}, 2018, pp. 7268--7277.

\bibitem{jiang2019integral}
P.-T. Jiang, Q.-B. Hou, Y.~Cao, M.-M. Cheng, Y.~Wei, and H.~Xiong, ``Integral
  object mining via online attention accumulation,'' in \emph{Proc. IEEE
  International Conference on Computer Vision}, 2019, pp. 2070--2079.

\bibitem{dai2015boxsup}
J.~Dai, K.~He, and J.~Sun, ``Boxsup: Exploiting bounding boxes to supervise
  convolutional networks for semantic segmentation,'' in \emph{Proc. IEEE
  International Conference on Computer Vision}, 2015, pp. 1635--1643.

\bibitem{khoreva2017simple}
A.~Khoreva, R.~Benenson, J.~Hosang, M.~Hein, and B.~Schiele, ``Simple does it:
  Weakly supervised instance and semantic segmentation,'' in \emph{Proc. IEEE
  Conference on Computer Vision and Pattern Recognition}, 2017, pp. 876--885.

\bibitem{song2019box}
C.~Song, Y.~Huang, W.~Ouyang, and L.~Wang, ``Box-driven class-wise region
  masking and filling rate guided loss for weakly supervised semantic
  segmentation,'' in \emph{Proc. IEEE Conference on Computer Vision and Pattern
  Recognition}, 2019, pp. 3136--3145.

\bibitem{lin2016scribblesup}
D.~Lin, J.~Dai, J.~Jia, K.~He, and J.~Sun, ``Scribblesup: Scribble-supervised
  convolutional networks for semantic segmentation,'' in \emph{Proc. IEEE
  Conference on Computer Vision and Pattern Recognition}, 2016, pp. 3159--3167.

\bibitem{vernaza2017learning}
P.~Vernaza and M.~Chandraker, ``Learning random-walk label propagation for
  weakly-supervised semantic segmentation,'' in \emph{Proc. IEEE Conference on
  Computer Vision and Pattern Recognition}, 2017, pp. 7158--7166.

\bibitem{bearman2016s}
A.~Bearman, O.~Russakovsky, V.~Ferrari, and L.~Fei-Fei, ``What's the point:
  Semantic segmentation with point supervision,'' in \emph{Proc. European
  Conference on Computer Vision}, 2016, pp. 549--565.

\bibitem{zhou2016learning}
B.~Zhou, A.~Khosla, {\`A}.~Lapedriza, A.~Oliva, and A.~Torralba, ``Learning
  deep features for discriminative localization,'' in \emph{Proc. IEEE
  Conference on Computer Vision and Pattern Recognition}, 2016, pp. 2921--2929.

\bibitem{selvaraju2017grad}
R.~R. Selvaraju, M.~Cogswell, A.~Das, R.~Vedantam, D.~Parikh, and D.~Batra,
  ``Grad-cam: Visual explanations from deep networks via gradient-based
  localization,'' in \emph{Proc. IEEE International Conference on Computer
  Vision}, 2017, pp. 618--626.

\bibitem{hou2017deeply}
Q.~Hou, M.-M. Cheng, X.~Hu, A.~Borji, Z.~Tu, and P.~Torr, ``Deeply supervised
  salient object detection with short connections,'' in \emph{Proc. IEEE
  Conference on Computer Vision and Pattern Recognition}, 2017, pp. 3203--3212.

\bibitem{lee2019ficklenet}
J.~Lee, E.~Kim, S.~Lee, J.~Lee, and S.~Yoon, ``Ficklenet: Weakly and
  semi-supervised semantic image segmentation using stochastic inference,'' in
  \emph{Proc. IEEE Conference on Computer Vision and Pattern Recognition},
  2019, pp. 5267--5276.

\bibitem{fan2020learning}
J.~Fan, Z.~Zhang, C.~Song, and T.~Tan, ``Learning integral objects with
  intra-class discriminator for weakly-supervised semantic segmentation,'' in
  \emph{Proc. IEEE Conference on Computer Vision and Pattern Recognition},
  2020, pp. 4283--4292.

\bibitem{fan2020employing}
J.~Fan, Z.~Zhang, and T.~Tan, ``Employing multi-estimations for
  weakly-supervised semantic segmentation,'' in \emph{Proc. European Conference
  on Computer Vision}, 2020, pp. 332--348.

\bibitem{sun2020mining}
G.~Sun, W.~Wang, J.~Dai, and L.~Gool, ``Mining cross-image semantics for weakly
  supervised semantic segmentation,'' in \emph{Proc. European Conference on
  Computer Vision}, 2020, pp. 347--365.

\bibitem{long2015fully}
J.~Long, E.~Shelhamer, and T.~Darrell, ``Fully convolutional networks for
  semantic segmentation,'' in \emph{Proc. IEEE Conference on Computer Vision
  and Pattern Recognition}, 2015, pp. 3431--3440.

\bibitem{ronneberger2015u}
O.~Ronneberger, P.~Fischer, and T.~Brox, ``U-net: Convolutional networks for
  biomedical image segmentation,'' in \emph{International Conference on Medical
  Image Computing and Computer Assisted Intervention}, 2015, pp. 234--241.

\bibitem{badrinarayanan2017segnet}
V.~Badrinarayanan, A.~Kendall, and R.~Cipolla, ``Segnet: A deep convolutional
  encoder-decoder architecture for image segmentation,'' vol.~39, no.~12, 2017,
  pp. 2481--2495.

\bibitem{chen2017deeplab}
L.-C. Chen, G.~Papandreou, I.~Kokkinos, K.~Murphy, and A.~Yuille, ``Deeplab:
  Semantic image segmentation with deep convolutional nets, atrous convolution,
  and fully connected crfs,'' vol.~40, no.~4, 2017, pp. 834--848.

\bibitem{zhao2017pyramid}
H.~Zhao, J.~Shi, X.~Qi, X.~Wang, and J.~Jia, ``Pyramid scene parsing network,''
  in \emph{Proc. IEEE Conference on Computer Vision and Pattern Recognition},
  2017, pp. 2881--2890.

\bibitem{zhang2018context}
H.~Zhang, K.~Dana, J.~Shi, Z.~Zhang, X.~Wang, A.~Tyagi, and A.~Agrawal,
  ``Context encoding for semantic segmentation,'' in \emph{Proc. IEEE
  Conference on Computer Vision and Pattern Recognition}, 2018, pp. 7151--7160.

\bibitem{liu2019auto}
C.~Liu, L.-C. Chen, F.~Schroff, H.~Adam, W.~Hua, A.~Yuille, and L.~Fei-Fei,
  ``Auto-deeplab: Hierarchical neural architecture search for semantic image
  segmentation,'' in \emph{Proc. IEEE Conference on Computer Vision and Pattern
  Recognition}, 2019, pp. 82--92.

\bibitem{xie2021scale}
G.-S. Xie, J.~Liu, H.~Xiong, and L.~Shao, ``Scale-aware graph neural network
  for few-shot semantic segmentation,'' in \emph{Proc. IEEE Conference on
  Computer Vision and Pattern Recognition}, 2021, pp. 5475--5484.

\bibitem{xie2021few}
G.-S. Xie, H.~Xiong, J.~Liu, Y.~Yao, and L.~Shao, ``Few-shot semantic
  segmentation with cyclic memory network,'' in \emph{Proc. IEEE International
  Conference on Computer Vision}, 2021, pp. 7293--7302.

\bibitem{chen2021semantically}
T.~Chen, G.~Xie, Y.~Yao, Q.~Wang, F.~Shen, Z.~Tang, and J.~Zhang,
  ``Semantically meaningful class prototype learning for one-shot image
  segmentation,'' vol.~24, pp. 968--980, 2022.

\bibitem{lin2017refinenet}
G.~Lin, A.~Milan, C.~Shen, and I.~Reid, ``Refinenet: Multi-path refinement
  networks for high-resolution semantic segmentation,'' in \emph{Proc. IEEE
  Conference on Computer Vision and Pattern Recognition}, 2017, pp. 1925--1934.

\bibitem{wang2018non}
X.~Wang, R.~Girshick, A.~Gupta, and K.~He, ``Non-local neural networks,'' in
  \emph{Proc. IEEE Conference on Computer Vision and Pattern Recognition},
  2018, pp. 7794--7803.

\bibitem{fu2019dual}
J.~Fu, J.~Liu, H.~Tian, Y.~Li, Y.~Bao, Z.~Fang, and H.~Lu, ``Dual attention
  network for scene segmentation,'' in \emph{Proc. IEEE Conference on Computer
  Vision and Pattern Recognition}, 2019, pp. 3146--3154.

\bibitem{huang2019ccnet}
Z.~Huang, X.~Wang, L.~Huang, C.~Huang, Y.~Wei, and W.~Liu, ``Ccnet: Criss-cross
  attention for semantic segmentation,'' in \emph{Proc. IEEE International
  Conference on Computer Vision}, 2019, pp. 603--612.

\bibitem{wang2021pyramid}
W.~Wang, E.~Xie, X.~Li, D.-P. Fan, K.~Song, D.~Liang, T.~Lu, P.~Luo, and
  L.~Shao, ``Pyramid vision transformer: A versatile backbone for dense
  prediction without convolutions,'' in \emph{Proc. IEEE International
  Conference on Computer Vision}, 2021, pp. 568--578.

\bibitem{wei2017object}
Y.~Wei, J.~Feng, X.~Liang, M.-M. Cheng, Y.~Zhao, and S.~Yan, ``Object region
  mining with adversarial erasing: A simple classification to semantic
  segmentation approach,'' in \emph{Proc. IEEE Conference on Computer Vision
  and Pattern Recognition}, 2017, pp. 1568--1576.

\bibitem{hou2018self}
Q.~Hou, P.-T. Jiang, Y.~Wei, and M.-M. Cheng, ``Self-erasing network for
  integral object attention,'' in \emph{Advances in Neural Information
  Processing Systems}, 2018, pp. 549--559.

\bibitem{Li2021GroupWiseSM}
X.~Li, T.~Zhou, J.~Li, Y.~Zhou, and Z.~Zhang, ``Group-wise semantic mining for
  weakly supervised semantic segmentation,'' in \emph{Proc. AAAI Conference on
  Artificial Intelligence}, 2021, pp. 1984--1992.

\bibitem{zhang2018adversarial}
X.~Zhang, Y.~Wei, J.~Feng, Y.~Yang, and T.~Huang, ``Adversarial complementary
  learning for weakly supervised object localization,'' in \emph{Proc. IEEE
  Conference on Computer Vision and Pattern Recognition}, 2018, pp. 1325--1334.

\bibitem{simonyan2014very}
K.~Simonyan and A.~Zisserman, ``Very deep convolutional networks for
  large-scale image recognition,'' in \emph{Proc. International Conference on
  Learning Representations}, 2015.

\bibitem{deng2009imagenet}
J.~Deng, W.~Dong, R.~Socher, L.-J. Li, K.~Li, and L.~Fei-Fei, ``Imagenet: A
  large-scale hierarchical image database,'' in \emph{Proc. IEEE Conference on
  Computer Vision and Pattern Recognition}, 2009, pp. 248--255.

\bibitem{chang2020weakly}
Y.-T. Chang, Q.~Wang, W.-C. Hung, R.~Piramuthu, Y.-H. Tsai, and M.-H. Yang,
  ``Weakly-supervised semantic segmentation via sub-category exploration,'' in
  \emph{Proc. IEEE Conference on Computer Vision and Pattern Recognition},
  2020, pp. 8991--9000.

\bibitem{zhang2020splitting}
T.~Zhang, G.~Lin, W.~Liu, J.~Cai, and A.~Kot, ``Splitting vs. merging: Mining
  object regions with discrepancy and intersection loss for weakly supervised
  semantic segmentation,'' in \emph{Proc. European Conference on Computer
  Vision}, 2020, pp. 663--679.

\bibitem{chen2020weakly}
L.~Chen, W.~Wu, C.~Fu, X.~Han, and Y.-T. Zhang, ``Weakly supervised semantic
  segmentation with boundary exploration,'' in \emph{Proc. European Conference
  on Computer Vision}, 2020, pp. 347--362.

\bibitem{zhang2020causal}
D.~Zhang, H.~Zhang, J.~Tang, X.-S. Hua, and Q.~Sun, ``Causal intervention for
  weakly-supervised semantic segmentation,'' in \emph{Advances in Neural
  Information Processing Systems}, vol.~33, 2020.

\bibitem{he2016deep}
K.~He, X.~Zhang, S.~Ren, and J.~Sun, ``Deep residual learning for image
  recognition,'' in \emph{Proc. IEEE Conference on Computer Vision and Pattern
  Recognition}, 2016, pp. 770--778.

\bibitem{hariharan2011semantic}
B.~Hariharan, P.~Arbel{\'a}ez, L.~D. Bourdev, S.~Maji, and J.~Malik, ``Semantic
  contours from inverse detectors,'' in \emph{Proc. IEEE International
  Conference on Computer Vision}, 2011, pp. 991--998.

\bibitem{wang2020weakly}
X.~Wang, S.~Liu, H.~Ma, and M.-H. Yang, ``Weakly-supervised semantic
  segmentation by iterative affinity learning,'' in \emph{International Journal
  of Computer Vision}, vol. 128, no.~6, 2020, pp. 1736--1749.

\bibitem{jin2017webly}
B.~Jin, M.~O. Segovia, and S.~S{\"u}sstrunk, ``Webly supervised semantic
  segmentation,'' in \emph{Proc. IEEE Conference on Computer Vision and Pattern
  Recognition}, 2017, pp. 3626--3635.

\bibitem{kim2017two}
D.~Kim, D.~Cho, and D.~Yoo, ``Two-phase learning for weakly supervised object
  localization,'' in \emph{Proc. IEEE International Conference on Computer
  Vision}, 2017, pp. 3534--3543.

\bibitem{li2018tell}
K.~Li, Z.~Wu, K.-C. Peng, J.~Ernst, and Y.~Fu, ``Tell me where to look: Guided
  attention inference network,'' in \emph{Proc. IEEE Conference on Computer
  Vision and Pattern Recognition}, 2018, pp. 9215--9223.

\bibitem{wang2018weakly}
X.~Wang, S.~You, X.~Li, and H.~Ma, ``Weakly-supervised semantic segmentation by
  iteratively mining common object features,'' in \emph{Proc. IEEE Conference
  on Computer Vision and Pattern Recognition}, 2018, pp. 1354--1362.

\bibitem{zeng2019joint}
Y.~Zeng, Y.~Zhuge, H.~Lu, and L.~Zhang, ``Joint learning of saliency detection
  and weakly supervised semantic segmentation,'' in \emph{Proc. IEEE
  International Conference on Computer Vision}, 2019, pp. 7223--7233.

\bibitem{sun2021ecs}
K.~Sun, H.~Shi, Z.~Zhang, and Y.~Huang, ``Ecs-net: Improving weakly supervised
  semantic segmentation by using connections between class activation maps,''
  in \emph{Proc. IEEE International Conference on Computer Vision}, 2021, pp.
  7283--7292.

\bibitem{saleh2016built}
F.~Saleh, M.~S. Aliakbarian, M.~Salzmann, L.~Petersson, S.~Gould, and J.~M.
  Alvarez, ``Built-in foreground/background prior for weakly-supervised
  semantic segmentation,'' in \emph{Proc. European Conference on Computer
  Vision}, 2016, pp. 413--432.

\bibitem{ahn2019weakly}
J.~Ahn, S.~Cho, and S.~Kwak, ``Weakly supervised learning of instance
  segmentation with inter-pixel relations,'' in \emph{Proc. IEEE Conference on
  Computer Vision and Pattern Recognition}, 2019, pp. 2209--2218.

\bibitem{shimoda2019self}
W.~Shimoda and K.~Yanai, ``Self-supervised difference detection for
  weakly-supervised semantic segmentation,'' in \emph{Proc. IEEE International
  Conference on Computer Vision}, 2019, pp. 5208--5217.

\bibitem{wang2020self}
Y.~Wang, J.~Zhang, M.~Kan, S.~Shan, and X.~Chen, ``Self-supervised equivariant
  attention mechanism for weakly supervised semantic segmentation,'' in
  \emph{Proc. IEEE Conference on Computer Vision and Pattern Recognition},
  2020, pp. 12\,275--12\,284.

\bibitem{liu2020leveraging}
Y.~Liu, Y.-H. Wu, P.-S. Wen, Y.-J. Shi, Y.~Qiu, and M.-M. Cheng, ``Leveraging
  instance-, image-and dataset-level information for weakly supervised instance
  segmentation,'' 2020.

\bibitem{yao2021non}
Y.~Yao, T.~Chen, G.-S. Xie, C.~Zhang, F.~Shen, Q.~Wu, Z.~Tang, and J.~Zhang,
  ``Non-salient region object mining for weakly supervised semantic
  segmentation,'' in \emph{Proc. IEEE Conference on Computer Vision and Pattern
  Recognition}, 2021, pp. 2623--2632.

\end{thebibliography}

\end{document}